%% file: peretroukhin_icra16.tex
\documentclass[letterpaper, 10 pt, conference]{ieeeconf}  

\IEEEoverridecommandlockouts
\pdfoutput=1

\usepackage{xparse} 
\usepackage{amsmath} 
\usepackage{amssymb}
\usepackage{amsfonts} 
\usepackage{bbm}

\usepackage{graphicx}
\usepackage{balance}
\usepackage{hyperref}
\usepackage{cleveref}

\usepackage{framed}

\usepackage{algpseudocode}
\usepackage{algorithm}

\newcommand\T{\rule{0pt}{2.6ex}}        

\usepackage{nomencl} 
\makenomenclature

\NewDocumentCommand\placeholder{mmo}{
  \framebox{
    \begin{minipage}[c][#2]{#1}
      \centering
      \IfNoValueTF{#3}{Placeholder}{#3}
    \end{minipage} 
  }
}

\input{variables}

\title{\LARGE \bf PROBE-GK: Predictive Robust Estimation using Generalized Kernels}

\author{Valentin Peretroukhin$^{1}$, William Vega-Brown$^{2}$, Nicholas Roy$^{2}$ and Jonathan Kelly$^{1}$\thanks{$^{1}${\footnotesize V. Peretroukhin and J. Kelly are with the Space \& Terrestrial Autonomous Robotic Systems Laboratory, Institute for Aerospace Studies, University of Toronto, {\tt v.peretroukhin@mail.utoronto.ca, jkelly@utias.utoronto.ca}}.}
\thanks{$^{2}${\footnotesize W. Vega-Brown and N. Roy are with the Robust Robotics Group, Computer Science and Artificial Intelligence Laboratory, Massachusetts Institute of Technology, {\tt \{wrvb,nickroy\}@csail.mit.edu}}.}
}

\begin{document} 

\maketitle 
\thispagestyle{empty}
\pagestyle{empty}

\begin{abstract}
  Many algorithms in computer vision and robotics make strong assumptions
  about uncertainty, and rely on the validity of these assumptions to produce
  accurate and consistent state estimates.  In practice, dynamic environments
  may degrade sensor performance in predictable ways that cannot be captured
  with static uncertainty parameters.  In this paper, we employ fast
  nonparametric Bayesian inference techniques to more accurately model sensor
  uncertainty.  By setting a prior on observation uncertainty, we derive a
  predictive robust estimator, and show how our model can be learned from sample
  images, both with and without knowledge of the motion used to generate the
  data.  We validate our approach through Monte Carlo simulations, and report
  significant improvements in localization accuracy relative to a fixed noise
  model in several settings, including on synthetic data, the KITTI dataset, and
  our own experimental platform.
\end{abstract}

\section{Introduction}
Modern ground, aerial, and underwater vehicles are able to carry exteroceptive
sensors capable of observing the world with high spatial and temporal
resolution.  Despite steady improvements in computing power, it remains
impractical in many situations for robots to reason directly over \textit{all}
of the available sensor data. Instead, it is common to use feature extraction
and interest point detection algorithms to provide a simplified representation
of the environment, and to perform tasks like odometry and mapping using that
simplified feature-based representation.

However, not all features are created equal; most feature-based methods rely on
random sample consensus algorithms \cite{fischler1981random} to partition the extracted features
into inliers and outliers, and perform estimation based only on inliers. It is
common to guard against misclassifying an outlier as an inlier by using robust
estimation techniques, such as the Cauchy costs employed in
\cite{kerl2013robust} or the dynamic covariance scaling devised
by \cite{Burgard:ii}. These approaches, often grouped under the title of M-estimation, aim to maintain a quadratic influence
of small errors, while reducing the contribution of larger errors. The
robustness and accuracy of feature-based visual odometry often hinges on the
tuning of the parameters of inlier selection and robust estimation. Performance
can vary significantly from one environment to the next, and most algorithms require
careful tuning to work in a given environment. 

\begin{figure}
    \centering
      \includegraphics[width=0.45\textwidth]{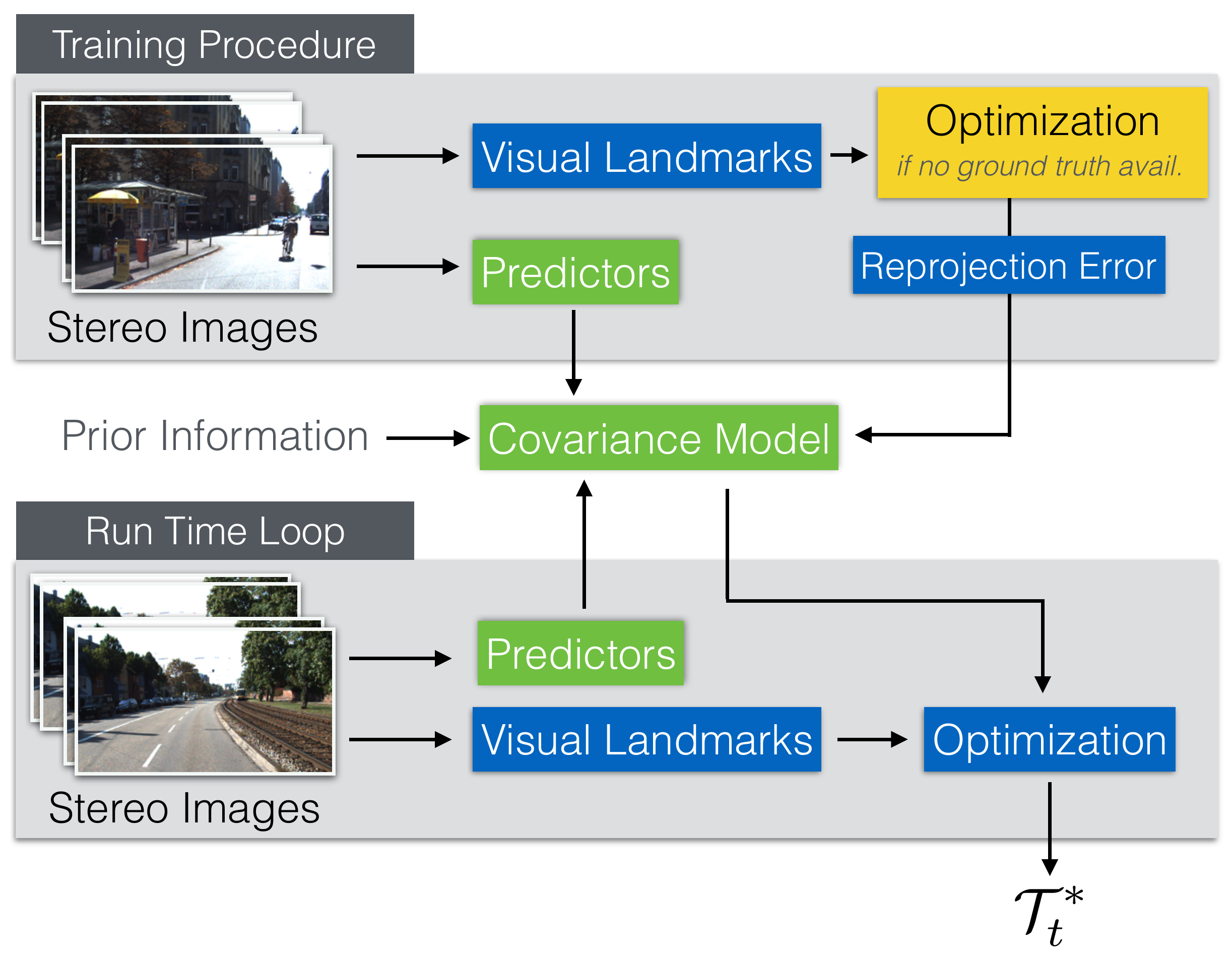}
      \caption{Our proposed system builds a predictive noise model for stereo visual odometry. (a)~At training
        time, we extract landmarks from two pairs of stereo images,
        and use egomotion ground truth to compute reprojection errors to build a covariance
        model. (b)~At run time, we predict a covariance for each visual landmark. We use these covariances in a robust nonlinear least-squares
        problem, which is solved to estimate the transform between camera poses.
        (c)~If the ground truth egomotion is not known, we iteratively apply an
        optimization procedure (yellow box) to estimate them.}
        \vspace{-1em}
    \label{fig:system}
\end{figure}

In this paper, we describe a principled, data-driven way to build a noise model
for visual odometry. We combine our previous work \cite{peretroukhin2015PROBE}
on predictive robust estimation (PROBE) with our work on covariance estimation
\cite{VegaBrown:2013fv} to formulate a predictive robust estimator for a
stereo visual odometry pipeline. We frame the traditional non-linear least
squares optimization problem as a problem of maximum likelihood estimation with
a Gaussian noise model, and infer a distribution over the covariance matrix of
the Gaussian noise from a predictive model learned from training data. This
results in a Student's~$t$ distribution over the noise, and naturally yields a
robust nonlinear least-squares optimization problem.  In this way, we can predict,
in a principled manner, how informative each visual feature is with respect to the final state
estimate, which allows our approach to intelligently weight observations to
produce more accurate odometry estimates.  Our pipeline is outlined in
Figure \ref{fig:system}.

The central contributions of our paper are: 
\begin{enumerate}
\item a probabilistic model for sparse stereo visual odometry, leading to
  a predictive robust algorithm for inference on that model,
\item a procedure for training our model using pairs of stereo images with
  known relative transform, and
\item an iterative, expectation-maximization approach to train our model when
  the relative ground truth egomotion is unavailable.
\end{enumerate}

\section{System Overview}
\subsection{Sparse stereo visual odometry}
\label{sec:ssvo}

In our frame-to-frame sparse stereo odometry pipeline,  the objective is to find
$\Transform_t\in\text{SE}(3)$, the rigid transform between two subsequent stereo camera poses (note that the temporal index $t$ refers to the set of two stereo camera poses). We begin by rectifying, then stereo and temporally
matching the set of 4 images to generate the corresponding locations of a set
of $N_t$ visual landmarks in each stereo pair.  Each landmark corresponds to a
point in space, expressed in homogeneous coordinates in the camera frame as
$\HomogeneousPoint{i}{t} := \Transpose{\bbm p_1 & p_2 & p_3 & p_4 \ebm} \in
\HomogeneousNumbers[3]$.  The stereo-camera model, $\ProjectionFunction$,
projects a landmark expressed in homogeneous coordinates into image space, so
that $\ImageLandmark{i}{t}$, the stereo pixel coordinates of landmark $i$ in the first camera pose at time $t$, is given
by 
\begin{equation}
	\ImageLandmark{i}{t} = \bbm u_l \\ v_l \\ u_r \\ v_r \ebm 
  = \ProjectionFunction(\HomogeneousPoint{i}{t}) 
  = \Matrix{M} \frac{1}{p_3}\HomogeneousPoint{i}{t},
\end{equation}
where
\begin{equation}
 \Matrix{M} = \bbm f_u & 0 & c_u & f_u \frac{b}{2} \\ 0 & f_v & c_v & 0 \\ f_u 
                        & 0 & c_u & -f_u\frac{b}{2} \\ 0 & f_b & c_v & 0 \ebm.
\end{equation}
Here, $\{c_u, c_v\}$, $\{f_u, f_v\}$, and $b$ are the principal points, focal
lengths and baseline of the stereo camera respectively. Note that in this
formulation, the stereo camera frame is centered between the two individual
lenses.  

We triangulate landmarks in the first camera frame, $\ImageLandmark{i}{t}$, and re-project
them into the second frame, $\ImageLandmark{i}{t}'$. We model errors due to sensor noise
and quantization as a Gaussian distribution in image space with a known covariance
$\Covariance$,
\begin{equation}
  p(\ImageLandmark{i}{t}' \vert \ImageLandmark{i}{t}, \Transform_t,
  \Covariance)
  =\NormalDistribution\left(\Vector{e}_{i,t}(\Transform_t); \Vector 0, \Covariance\right), 
\end{equation}
where
\begin{equation}
 \Vector{e}_{i,t} = \ImageLandmark{i}{t}' - \ProjectionFunction( \Transform_t 
    \ProjectionFunction^{-1}( \ImageLandmark{i}{t} ) ).	
   \label{eq:image_error}
\end{equation}
  The maximum likelihood transform,
$\Transform_t^*$, is then given by 
\begin{equation}
  \Transform_t^* = \ArgMin{\Transform_t\in\text{SE}(3)}\sum_{i=1}^{N_t} 
  \Transpose{\Vector{e}_{i,t}} \Covariance^{-1} \Vector{e}_{i,t}.
\end{equation}
This is a nonlinear least squares problem, and can be solved iteratively using
standard techniques. During iteration $n$, we represent the transform as the
product of an estimate $\Transform^{(n)}\in\text{SE}(3)$ and a perturbation
$\delta\Vector{\xi}\in\RealNumbers[6]$ represented in exponential
coordinates:
\begin{equation}
  \Transform_t = \exp{\left( \delta\Vector{\xi}^{\wedge}
  \right)} \Transform_t^{(n)}.
\end{equation}
The wedge operator $(\cdot)^\wedge$ is defined (following
\cite{barfoot2014associating}) as both the map
$\RealNumbers[3]\to\mathfrak{so}(3)$,  
\begin{equation}
\Vector \phi ^\wedge \triangleq \bbm \phi_1 \\ \phi_2 \\ \phi_3 \ebm^\wedge	= \bbm 0 &
-\phi_3 & \phi_2 \\ \phi_3 & 0 & -\phi_1 \\ -\phi_2 & \phi_1 & 0 \ebm,
\label{eq:wedgeOpRotation}
\end{equation}
and the map $\RealNumbers[6]\to\mathfrak{se}(3)$,
\begin{equation}
  \Vector \xi^\wedge \triangleq \bbm \Vector \rho \\ \Vector \phi \ebm ^\wedge = \bbm
  \Vector \phi^\wedge & \Vector \rho \\ \Transpose{\Vector{0}} &  0 \ebm.	
\end{equation}
Linearizing the transform for small perturbations $\delta\Vector{\xi}$
yields a linear least-squares problem:
\begin{equation}
  \mathcal{L}(\delta \Vector{\xi}) = \frac{1}{2}\sum_{i=1}^{N_t} 
  \Transpose{\left(\Vector{e}_{i,t}^{(n)}
  - \Matrix J_{i,t}^{(n)} \delta\Vector{\xi}\right)}
\Covariance^{-1}
 \left(\Vector{e}_{i,t}^{(n)}
 - \Matrix J_{i,t}^{(n)} \delta\Vector{\xi}\right)
  \end{equation}
Here, $\Matrix J_{i,t}^{(n)}$ is the Jacobian matrix of the reprojection error.
The explicit form of the Jacobian matrix is omitted for brevity but can be
found in our supplemental materials.\footnote{\url{http://groups.csail.mit.edu/rrg/peretroukhin_icra16/supplemental.pdf}}

Rearranging, we see the minimizing perturbation is the solution to a
linear system of equations:
\begin{equation}
  \delta\Vector{\xi}^{(n)} = 
  \left( \sum_{i=1}^{N_t} \Transpose{\Matrix J}_{i,t}
  \ImageLandmarkCovariance{}{}^{-1} \Matrix J_{i,t} \right)^{-1}
  \sum_{i=1}^{N_t} \Transpose{\Matrix J}_{i,t}
  \Covariance^{-1} \Vector{e}_{i,t}^{(n)}. 
\label{eq:least-squares-iteration}
\end{equation}
We then update the estimated transform and proceed to the next iteration.
\begin{equation}
  \Transform_t^{(n+1)} = \exp{\left( \delta\Vector{\xi}^{(n)\wedge}
  \right)} \Transform_t^{(n)}. \label{eq:update}
\end{equation}
There are many reasonable choices for both the initial transform
$\Transform_t^{(0)}$ and for the conditions under which we terminate
iteration. We initialize the estimated transform to identity, and iteratively
perform the update given by \cref{eq:update} until we see a relative change in
the squared error of less than one percent after an update.  

\subsection{Predictive noise models for visual odometry}

The process described in the previous section employs a fixed noise covariance
$\Covariance$.  However, not all landmarks are created equal: differing texture
gradients can cause feature localization to degrade in predictable ways, and
effects like motion blur can lead to landmarks being less informative. If we had a
good estimate of the noise covariance for each landmark, we could simply replace the fixed covariance $\Covariance$
with one that varies for each stereo observation,
$\ImageLandmarkCovariance{i}{t}$. Such a predictive model would allows us to  better account for observation errors from a diverse set of noise sources, and incorporate information from landmarks that may otherwise be discarded by a binary outlier rejection scheme.

However, estimating these covariances in a principled way is a nontrivial task.
Even when we have reasonable heuristic estimates available, it is difficult
to guarantee those estimates will be reliable.  Instead of relying solely on such
heuristics, we propose to learn these image-space noise covariances from data.

We associate with each landmark $\ImageLandmark{i}{t}$ a vector of
\emph{predictors}, $\Predictor{i}{t}\in\RealNumbers[M]$.  Each predictor can be computed using both visual and inertial cues, allowing us to model effects like motion blur and self-similar
textures. We then compute the covariance as a function of these predictors, so
that $\ImageLandmarkCovariance{i}{t} = \Covariance( \Predictor{i}{t})$.  In
order to exploit conjugacy to a Gaussian noise model, we formulate our prior knowledge
about this function using an inverse Wishart (IW) distribution over positive
definite $d \times d$ matrices (the IW distribution has been used as a prior on covariance matrices in other robotics and computer vision contexts, see for example, \cite{fitzgibbon2007learning}). This distribution is defined by a scale matrix
$\Matrix{\Psi} \in \RealNumbers[d][d]\succ0$ and a scalar quantity called the
degrees of freedom $\nu\in\RealNumbers>d-1$:
\begin{align}
  p\left(\Matrix{R}\right) &= \InverseWishartDistribution
  \left(\Matrix{R}; \Matrix{\Psi}, \nu\right)  \\ 
  &= \frac{\vert\Matrix{\Psi}\vert^{\nu/2}}{2^{\frac{\nu
  d}{2}}\Gamma_d(\frac{\nu}{2})} \vert\Matrix{R}\vert
  ^{-\frac{\nu+d+1}{2}} \exp\left( -\frac{1}{2} \text{tr}\left(\Matrix{\Psi}
  \Matrix{R}^{-1}\right)\right) \nonumber.
\end{align}
We use the scale matrix to encode our prior estimate of the
covariance, and the degrees of freedom to encode our confidence in
that estimate.  Specifically, if we estimate the covariance $\Matrix{R}$
associated with predictor $\Vector{\phi}$ to be $\hat{\Matrix{R}}$ with a
confidence equivalent to seeing $n$ independent samples of the error from
$\NormalDistribution(\Vector{0}, \hat{\Matrix R})$, we would choose
$\nu(\Vector{\phi})=n$ and $\Matrix{\Psi}(\Vector{\phi})=n\hat{\Matrix{R}}$.

Given a sequence of observations and ground truth transformations,
\begin{equation}
\mathcal{D}=\{\mathcal{I}_t,\Transform_t\},\quad t\in[1,N]
\end{equation} 
where
\begin{equation}
 \mathcal{I}_t = \{\ImageLandmark{i}{t},
\ImageLandmark{i}{t}', \Predictor{i}{t} \} \quad i\in[1,N_t],
\end{equation}
we can use the procedure of generalized kernel estimation
\cite{vega-brown2014nonparametric} to infer a posterior distribution over the
covariance matrix $\TargetImageLandmarkCovariance$ associated with some query
predictor vector $\TargetPredictor$:
\begin{align}
  p(\TargetImageLandmarkCovariance|\mathcal{D}, \TargetPredictor) &\propto
    \prod_{i,t}\mathcal{N}(\Vector{e}_{i,t} \vert \Vector{0},
      \TargetImageLandmarkCovariance)^{k(\TargetPredictor,\Predictor{i}{t})} \nonumber
      \\
      &\qquad\times\text{IW}(\TargetImageLandmarkCovariance;\Matrix{\Psi}(\TargetPredictor),
      \nu(\TargetPredictor)) \\
      &=\text{IW}(\TargetImageLandmarkCovariance;\Matrix{\Psi}_*, \nu_*). 
\end{align}
Here, $\Vector{e}_{i,t}= \ImageLandmark{i}{t}' - \ProjectionFunction(
\Transform_t \ProjectionFunction^{-1}( \ImageLandmark{i}{t} ))$ as before.  
The function $k: \RealNumbers[M]\times\RealNumbers[M]\to[0,1]$ is a kernel
function which measures the similarity of two points in predictor space.
Note also that the posterior parameters $\Matrix{\Psi}_*$ and $\nu_*$ can be
computed in closed form as
\begin{align}
  \Matrix{\Psi}_* &= \Matrix{\Psi}(\TargetPredictor) + 
    \sum_{i,t} k(\TargetPredictor,\Predictor{i}{t}) 
    \Vector{e}_{i,t}\Transpose{\Vector{e}_{i,t}}, \label{eq:compute-psi}\\
  \nu_* &= \nu(\TargetPredictor) + \sum_{i,t}
    k(\TargetPredictor,\Predictor{i}{t}).  \label{eq:compute-nu}
\end{align}

If we marginalize over the covariance matrix, we find that the posterior
predictive distribution is a multivariate Student's~$t$ distribution:
\begin{align}
p(\ImageLandmark{i}{t}'&|  \Transform_t,\ImageLandmark{i}{t}, \mathcal{D},
  \Predictor{i}{t}) \\ &= \int \mathrm{d}\ImageLandmarkCovariance{i}{t}
  \NormalDistribution\left( \Vector{e}_{i,t}; \Vector{0},
    \ImageLandmarkCovariance{i}{t}\right)
  \text{IW}(\ImageLandmarkCovariance{i}{t};\Matrix{\Psi}_*, \nu_*)  \\ &=
  \StudentTDistribution_{\nu_*-d+1}\left(
    \Vector{e}_{i,t}; \Vector{0}, \frac{1}{\nu_*-d+1}\Matrix{\Psi}_*\right) \\ &=
    \frac{\Gamma(\frac{\nu_*+1}{2})}{\Gamma(\frac{\nu_*-d+1}{2})}
    \vert\Matrix{\Psi_*}\vert^{-\frac{1}{2}}\pi^{-\frac{d}{2}} \left(1+
    \Transpose{\Vector{e}_{i,t}} \Matrix{\Psi_*}^{-1} \Vector{e}_{i,t}
  \right)^{-\frac{\nu_*+1}{2}}. 
\end{align}
Given a new landmark and predictor vector, we can infer a noise model by
evaluating \cref{eq:compute-psi,eq:compute-nu}.  In order to accelerate this
computation, it is helpful to choose a kernel function with finite support:
that is, $k(\Vector{\phi},\Vector{\phi}')=0$ if $\Vert\Vector{\phi} -
\Vector{\phi}'\Vert_2 > \rho$. Then, by indexing our training data in a spatial
index such as a $k$-d tree, we can identify the subset of samples relevant to
evaluating the sums in \cref{eq:compute-psi,eq:compute-nu} in $\mathcal{O}(\log
N + \log N_t)$ time.  \Cref{alg:train-ground-truth} describes the procedure for
building this model. 

\begin{algorithm}
  \caption{Build the covariance model given a sequence of observations, $\mathcal{D}$.}
  \label{alg:train-ground-truth}
  \begin{algorithmic}
    \Function{BuildCovarianceModel}{$\mathcal{D}$}
      \State Initialize an empty spatial index $\mathcal{M}$
      \ForAll{$\mathcal{I}_t,\Transform_t$ in $\mathcal{D}$}
        \ForAll{$\{\ImageLandmark{i}{t}, \ImageLandmark{i}{t}',
          \Predictor{i}{t} \}$ in $\mathcal{I}_t$}
          \State $\Vector{e}_{i,t} = \ImageLandmark{i}{t}' -
            \ProjectionFunction( \Transform_t \ProjectionFunction^{-1}(
            \ImageLandmark{i}{t} ))$
            \State Insert $\Predictor{i}{t}$ into $\mathcal{M}$ and store
              $\Vector{e}_{i,t}$ at its location
        \EndFor
      \EndFor
      \State\Return $\mathcal{M}$
    \EndFunction
  \end{algorithmic}
\end{algorithm}

Once we have inferred a noise model for each landmark in a new image pair, the
maximum likelihood optimization problem is given by 
\begin{equation}
  \Transform_t^* = \ArgMin{\Transform_t\in\text{SE}(3)}\sum_{i=1}^{N_t} 
  (\nu_{i,t}+1)\log \left(1+ \Transpose{\Vector{e}_{i,t}}
  \Matrix{\Psi}_{i,t}^{-1} \Vector{e}_{i,t} \right).\label{eq:robust-loss}
\end{equation}
The final optimization problem thus emerges as a nonlinear least squares problem with a rescaled Cauchy-like loss
function, with error term $\Transpose{\Vector{e}_{i,t}}
(\frac{1}{\nu_{i,t}+1}\Matrix{\Psi}_{i,t})^{-1} \Vector{e}_{i,t}$ and outlier
scale $\nu_{i,t}+1$.  This is a common robust loss function which is
approximately quadratic in the reprojection error for
$\Transpose{\Vector{e}_{i,t}} \Matrix{\Psi}_{i,t}^{-1} \Vector{e}_{i,t} \ll
\nu_{i,t}+1$, but grows only logarithmically for $\Transpose{\Vector{e}_{i,t}}
\Matrix{\Psi}_{i,t}^{-1} \Vector{e}_{i,t} \gg \nu_{i,t}+1$.  It follows that in
the limit of large $\nu_{i,t}$---in regions of predictor space where there are
many relevant samples---our optimization problem becomes the original
least-squares optimization problem.

Solving nonlinear optimization problems with the form of  \Cref{eq:robust-loss}
is a well-studied and well-understood task, and software packages to
perform this computation are readily available. 
\Cref{alg:compute-transform} describes the procedure for computing the transform
between a new image pair, treating the optimization of \Cref{eq:robust-loss} as
a subroutine.

\begin{algorithm}
  \caption{Compute the transform between two images, given a set, $\mathcal{I}_t$,
    of landmarks and predictors extracted from an image pair and a covariance
    model $\mathcal{M}$. }
  \label{alg:compute-transform}
  \begin{algorithmic}
    \Function{ComputeTransform}{$\mathcal{I}_t$, $\mathcal{M}$}
      \ForAll{$\{\ImageLandmark{i}{t}, \ImageLandmark{i}{t}',
      \Predictor{i}{t} \}$ in $\mathcal{I}_t$}
        \State $\Matrix{\Psi}, \nu\gets$ \Call{InferNoiseModel} {$\mathcal{M}$, $
          \Predictor{i}{t}$}
        \State $g(\Transform) = \ImageLandmark{i}{t} -
          \ProjectionFunction( \Transform\ProjectionFunction^{-1}(
          \ImageLandmark{i}{t}' ))$
        \State $\mathcal{L} \gets \mathcal{L} +
        (\nu+1)\log\left(1 + \Transpose{g(\Transform)}
          \Matrix{\Psi}^{-1}
        g(\Transform)\right)$
      \EndFor
      \State \Return $\ArgMin{\Transform\in\text{SE}(3)}\mathcal{L}(\Transform)$
    \EndFunction
    \Function{InferNoiseModel}{$\mathcal{M}$, $\TargetPredictor$}
      \State $\textsc{neighbors}\gets$ \Call{GetNeighbors}{$\mathcal{M},
          \TargetPredictor, \rho$}
      \State\Comment{$\rho$ is the radius of the support of the kernel $k$}
      \State $\Matrix{\Psi}_* \gets \Matrix{\Psi}(\TargetPredictor)$ \State
      $\nu_* \gets \nu(\TargetPredictor)$
      \For{$(\Predictor{i}{t},\Vector{e}_{i,t})$ in $\textsc{Neighbors}$}
        \State $\Matrix{\Psi}_* \gets \Matrix{\Psi}_* + k(\TargetPredictor,
          \Predictor{i}{t}) \Vector{e}_{i,t}\Transpose{\Vector{e}_{i,t}}$
        \State $\nu_* \gets \nu_* + k(\TargetPredictor, \Predictor{i}{t})$
      \EndFor
    \State \Return $\Matrix{\Psi}_*, \nu_*$
    \EndFunction
  \end{algorithmic}
\end{algorithm}

We observe that \Cref{alg:compute-transform} is predictively robust, in the
sense that it uses past experiences not just to predict the reliability of a
given image landmark, but also to introspect and estimate its own knowledge of
that reliability.  Landmarks which are not known to be reliable are trusted 
less than landmarks which look like those which have been observed previously, where ``looks like'' is defined by our prediction space and choice of kernel. 

\subsection{Inference without ground truth}

\Cref{alg:train-ground-truth} requires access to the true transform between
training image pairs.  In practice, such ground truth data may be difficult
to obtain.  In these cases, we can instead formulate a likelihood model $p(\mathcal{D}' \vert
\Transform_1, \dots, \Transform_t)$, where $\mathcal{D}' = \{\mathcal{I}_t\}$ is a dataset consisting only of
landmarks and predictors for each training image pair. We can construct a model
for future queries by inferring the most likely sequence of transforms for our
training images.  The likelihood has the following factorized form:
\begin{multline}
  p(\mathcal{D}' \vert \Transform_{1:T}) \propto \int \prod_{i,t}
  \mathrm{d}\ImageLandmarkCovariance{i}{t}\,
  p(\ImageLandmark{i}{t}' \vert \ImageLandmark{i}{t}, 
    \Transform_t, \ImageLandmarkCovariance{i}{t}) \nonumber \\
  \times p(\ImageLandmarkCovariance{i}{t}\vert \Predictor{i}{t},
    \mathcal{D}, \Transform_{1:T}).
\end{multline}
We cannot easily maximize this likelihood, since marginalizing over the
noise covariances removes the independence of the transforms between
each image pair. To render the optimization tractable, we follow
our previous work \cite{VegaBrown:2013fv} and formulate an iterative expectation-maximization (EM)
procedure. Given an estimate $\Transform_{t}^{(n)}$ of the transforms, we can
compute the expected log-likelihood conditioned on our current estimate: 
\begin{multline}
  Q(\Transform_{1:T} \vert \Transform_{1:T}^{(n)}) = 
    \int \left(\prod_{i,t}\mathrm{d}\ImageLandmarkCovariance{i}{t}
      \,p(\ImageLandmarkCovariance{i}{t} | \mathcal{D}_{\setminus
        i,t},
      \Transform_{1:T}^{(n)})\right)
    \\ \times \log \prod_{i,t} p(\ImageLandmark{i}{t}' \vert
      \ImageLandmark{i}{t}, \Transform_t,
      \ImageLandmarkCovariance{i}{t}).
\end{multline}
This has the effect of rendering the likelihood of each transform to be
estimated independently.  Moreover, the expected log-likelihood can be
evaluated in closed form:
\begin{align}
  Q(\Transform_{1:T} | \Transform_{1:T}^{(n)}) \cong -\frac{1}{2}\sum_{t=1}^T
  \sum_{i=1}^{N_t} \Transpose{\Vector{e}_{i,t}}
  \left(\frac{1}{\nu_{i,t}^{(n)}}\Matrix{\Psi}_{i,t}^{(n)}\right)^{-1}
  \Vector{e}_{i,t}.
\end{align}
The symbol $\cong$ is used to indicate equality up to an additive constant. 
A derivation of this observation can be found in our supplemental material.

We can iteratively refine our estimate by maximizing the expected
log-likelihood
\begin{equation}
  \Transform_{1:T}^{(n+1)} = 
    \ArgMax {\Transform_{1:T}\in\text{SE}(3)^T}
    Q(\Transform_{1:T} \vert \Transform_{1:T}^{(n)}).
\end{equation}
Due to the additive structure of $Q(\Transform_{1:T} \vert
\Transform_{1:T}^{(n)})$, this takes the form of $T$ separate nonlinear least-squares optimizations:  
\begin{equation}
\label{eq:Qargmin}
  \Transform_{t}^{(n+1)} = 
    \ArgMin {\Transform_{t}\in\text{SE}(3)}
  \sum_{i=1}^{N_t} \Transpose{\Vector{e}_{i,t}}
  \left(\frac{1}{\nu_{i,t}^{(n)}}\Matrix{\Psi}_{i,t}^{(n)}\right)^{-1}
  \Vector{e}_{i,t}.
\end{equation}
 \Cref{alg:train-em} describes the process of training a model without
ground truth. We refer to this process as PROBE-GK-EM, and distinguish it from
PROBE-GK-GT (Ground Truth). We note that the sequence of estimated transforms,
$\Transform_{1:T}^{(n)}$, is guaranteed to converge to a local maxima of the
likelihood function \cite{dempster1977maximum}. It is also possible to use a robust loss function (\Cref{eq:robust-loss}) in place of \Cref{eq:Qargmin} during EM training. Although not formally motivated by the derivation  above, this approach often leads to lower test errors in practice. Characterizing when and why this robust learning process outperforms its non-robust alternative is part of ongoing work.

\begin{algorithm}
  \caption{Build the covariance model without ground truth given a sequence of observations, $\mathcal{D'}$, and an initial odometry estimate $\Transform_{1:T}^{(0)}$.}
  \label{alg:train-em}
  \begin{algorithmic}
    \Function{BuildCovarianceModel}{$\mathcal{D'}$, $\Transform_{1:T}^{(0)}$}
      \State Initialize an empty spatial index $\mathcal{M}$
      \ForAll{$\mathcal{I}_t$ in $\mathcal{D'}$}
        \ForAll{$\{\ImageLandmark{i}{t}, \ImageLandmark{i}{t}',
        \Predictor{i}{t} \}$ in $\mathcal{I}_t$}
          \State $\Vector{e}_{i,t} = \ImageLandmark{i}{t} -
          \ProjectionFunction( \Transform_t^{(0)} \ProjectionFunction^{-1}(
            \ImageLandmark{i}{t}' ))$
            \State Insert $\Predictor{i}{t}$ into $\mathcal{M}$ and store
              $\Vector{e}_{i,t}$ at its location
        \EndFor
      \EndFor
      \Repeat
        \ForAll{$\mathcal{I}_t$ in $\mathcal{D'}$}
          \ForAll{$\{\ImageLandmark{i}{t}, \ImageLandmark{i}{t}',
          \Predictor{i}{t} \}$ in $\mathcal{I}_t$}
            \State $\Matrix{\Psi}, \nu\gets$ \Call{InferNoiseModel} {$\mathcal{M}, \Predictor{i}{t}$}
            \State $g(\Transform) = \ImageLandmark{i}{t} -
              \ProjectionFunction( \Transform\ProjectionFunction^{-1}(
              \ImageLandmark{i}{t}' ))$
            \State $\mathcal{L} \gets \mathcal{L} +
              \Transpose{g(\Transform)}
              \left(\frac{1}{\nu}\Matrix{\Psi}\right)^{-1}
              g(\Transform)$
          \EndFor
          \State $\Transform_t \gets \ArgMin{\Transform\in\text{SE}(3)}
            \mathcal{L}(\Transform)$
          \State $\Vector{e}_{i,t} = \ImageLandmark{i}{t} -
            \ProjectionFunction( \Transform_t^{(0)} \ProjectionFunction^{-1}(
            \ImageLandmark{i}{t}' ))$
          \State Update the error stored at $\Predictor{i}{t}$ in $\mathcal{M}$
            to $\Vector{e}_{i,t}$
        \EndFor
      \Until{converged}
      \State\Return $\mathcal{M}$
    \EndFunction
  \end{algorithmic}
\end{algorithm}

\subsection{Implementation Details}
We implemented PROBE-GK using a combination of MATLAB and C++. We used the
open-source library \texttt{LIBVISO2} \cite{geiger2011stereoscan} for feature
extraction and matching. We implemented our own Levenberg-Marquardt optimization
routine, and used a custom C++ library to maintain the covariance model and
perform inference. 

\section{Results and Discussion}
To validate PROBE-GK, we used three types of data: synthetic simulations, the
KITTI dataset, and our own experimental data collected at the University of
Toronto. 

\subsection{Simulation}
\subsubsection{Monte-Carlo Verification}
\begin{figure}
    \centering
      \includegraphics[width=0.45\textwidth]{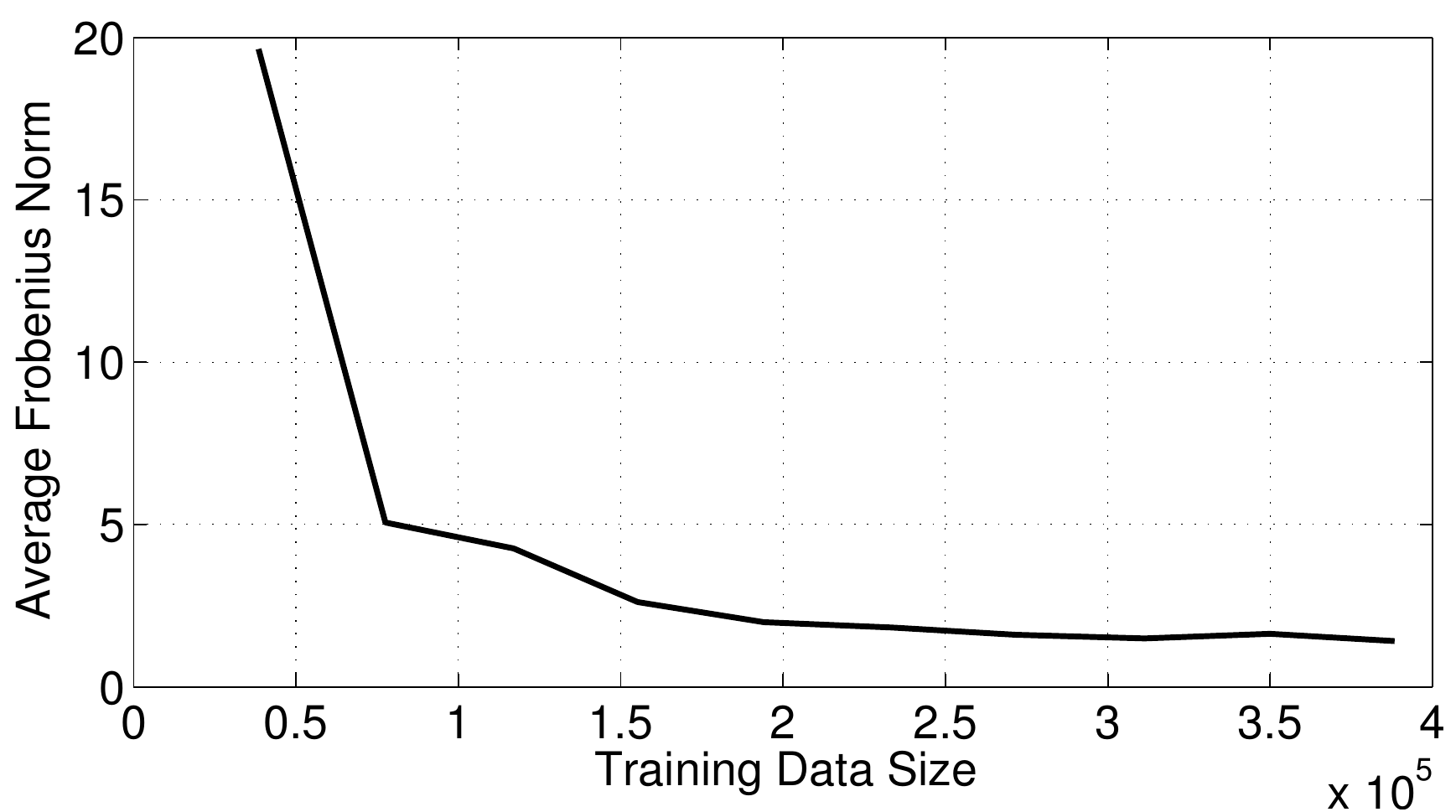}
     \caption{Mean Frobenius norm of the error between the estimated and true
       noise covariance as a function of training data size. The norm tends to
       zero as training data is added which indicates that PROBE-GK is learning
       the correct covariances.}
    \label{fig:frobNorm}
\end{figure}

To begin, we verified that PROBE-GK can predict increasingly accurate estimates of the true error covariance as more training data is added. We developed a basic simulation environment consisting of a large amount of point landmarks being observed by a stereo camera. In our simulation, the camera traversed a single step in one direction, and recorded  empirical reprojection errors based on ground truth poses. We simulated additive Gaussian noise on image coordinates, and used Monte Carlo simulations (propagating the additive noise through \Cref{eq:image_error}) to estimate the true covariances. \Cref{fig:frobNorm} shows the mean Frobenius norm (as defined in \cite{barfoot2014associating}) between the covariances estimated by PROBE-GK and the true covariances for a test trial. The mean norm tends to zero as more landmarks are added, indicating that PROBE-GK does learn the correct covariances.

\subsubsection{Synthetic World}
Next, we formulated a synthetic dataset wherein a stereo camera traverses a circular path observing 2000 randomly distributed point features.
We added Gaussian noise to each of the ideal projected
pixel co-ordinates for visible landmarks at every step. We varied the noise variance as a function of the vertical pixel coordinate of
the feature in image space. In addition, a small subset of the landmarks received an error term drawn from a uniform distribution to simulate the presence of outliers. The prediction space was composed of
the vertical and horizontal pixel locations in each of the stereo cameras.

\begin{figure}
    \centering
    \includegraphics[width=0.45\textwidth]{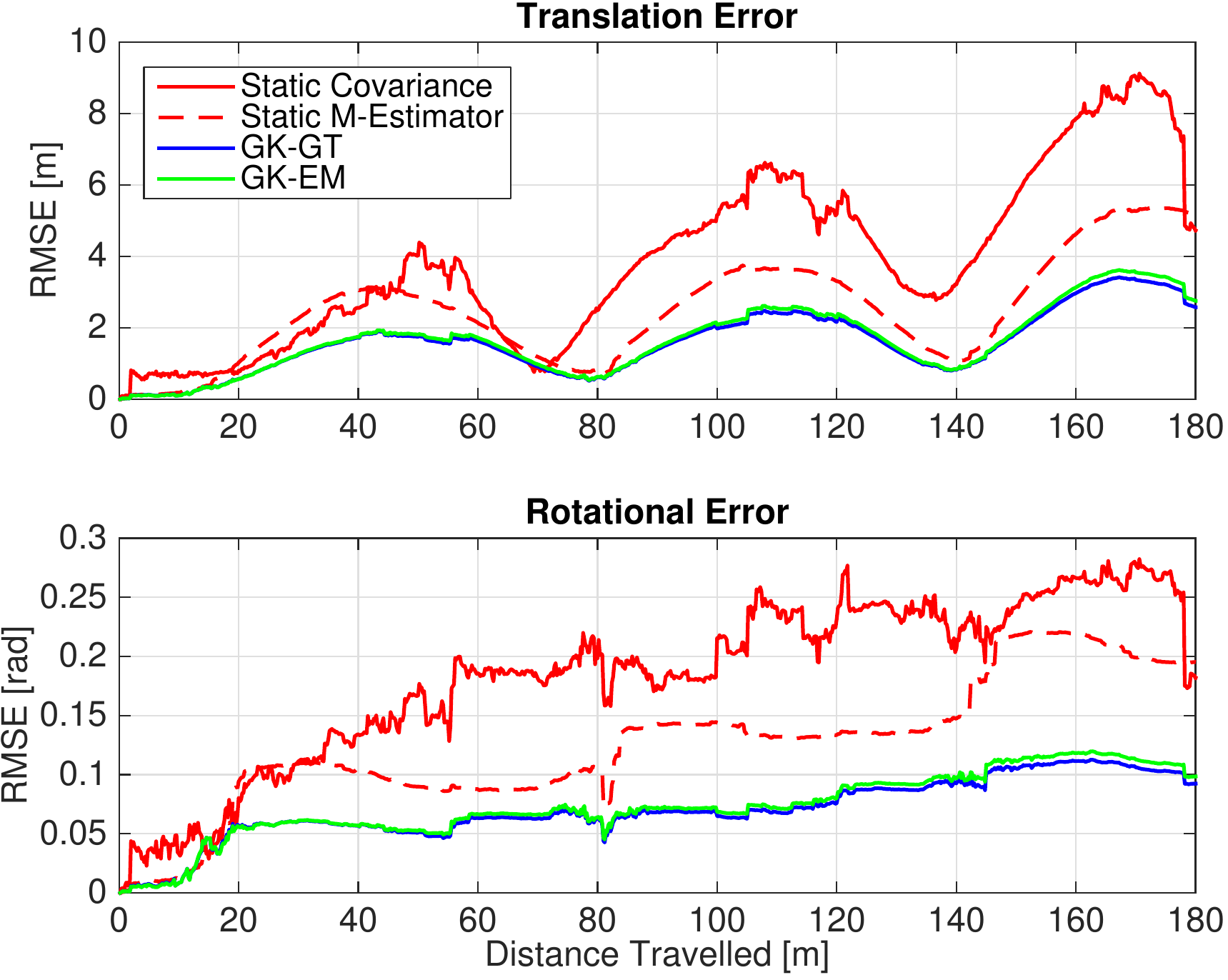}
    \caption{A comparison of translational and rotational Root Mean Square Error on simulated data
    (RMSE) for four different stereo-visual odometry pipelines: two baseline
    bundle adjustment procedures with and without a robust Student's~$t$ cost with a fixed and
    hand-tuned covariance and degrees of freedom (M-Estimation), a robust bundle
    adjustment with covariances learned from ground truth with
    \cref{alg:train-ground-truth} (GK-GT), and a robust bundle adjustment using
    covariances learned without ground truth using expectation maximization,
    with \cref{alg:train-em} (GK-EM). Note in this experiment, the RMSE curves
    for GK-GT and GK-EM very nearly overlap. The overall translational and
    rotational ARMSE values are shown in Table \ref{table:armse_errors}.} 
    \label{fig:sim_comparison}
\end{figure}

We simulated independent training and test traversals, where the camera moved for
30 and 60 seconds respectively (at a forward speed of 3 metres per second for final path
lengths of 90 and 180 meters).  \Cref{fig:sim_comparison} and \Cref{table:armse_errors} document the qualitative and quantitative comparisons of PROBE-GK (trained with and without ground-truth) against two baseline stereo odometry frameworks. Both baseline estimators were implemented based on \Cref{sec:ssvo}. The first utilized fixed covariances for all reprojection errors, while the second used a modified robust cost (i.e. M-estimation) based on Student's~$t$ weighting, with $\nu = 5$ (as suggested in \cite{kerl2013robust}).  These benchmarks served as baseline estimators (with and without robust costs) that used fixed covariance matrices and did not include a predictive component. 

Using PROBE-GK with ground truth data for training,
we significantly reduced both the translation and rotational Average Root Mean Squared Error (ARMSE)
by approximately 50\%. In our synthetic data, the Expectation Maximization approach was able to achieve nearly identical results to the ground-truth-aided model within 5 iterations.  

\subsection{KITTI Dataset}

\begin{figure*}
    \centering
    \includegraphics[width=0.98\textwidth]{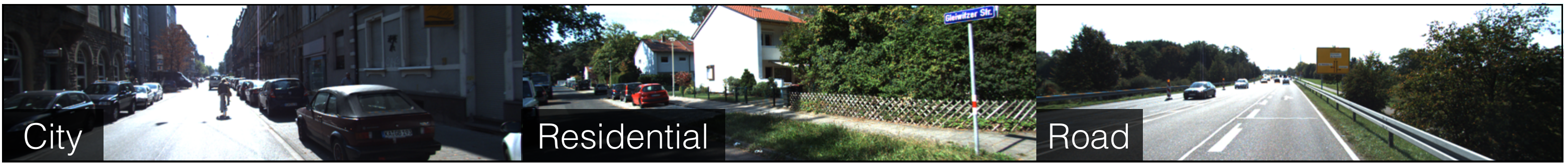}
    \caption{The KITTI dataset contains three different environments. We
      validate PROBE-GK by training on each type and testing against a baseline
      stereo visual odometry pipeline.}
      \vspace{-0.5em}
    \label{fig:kitti_environments}
\end{figure*}

To evaluate PROBE-GK on real environments, we trained and tested several models on
the KITTI Vision Benchmark suite \cite{geiger2012kitti, geiger2013vision}, a series of datasets collected by a car outfitted with a number of
sensors driven around different parts of Karlsruhe, Germany. Within the dataset, ground truth pose information is
provided by a high grade inertial navigation unit which also fuses measurements from differential GPS. Raw data
is available for different types of environments through which the car was driving; for our
work, we focused on the city, residential and road categories
(\Cref{fig:kitti_environments}).  From each category, we chose two separate trials for training and testing.

\begin{figure}
    \centering
    \includegraphics[width=0.45\textwidth]{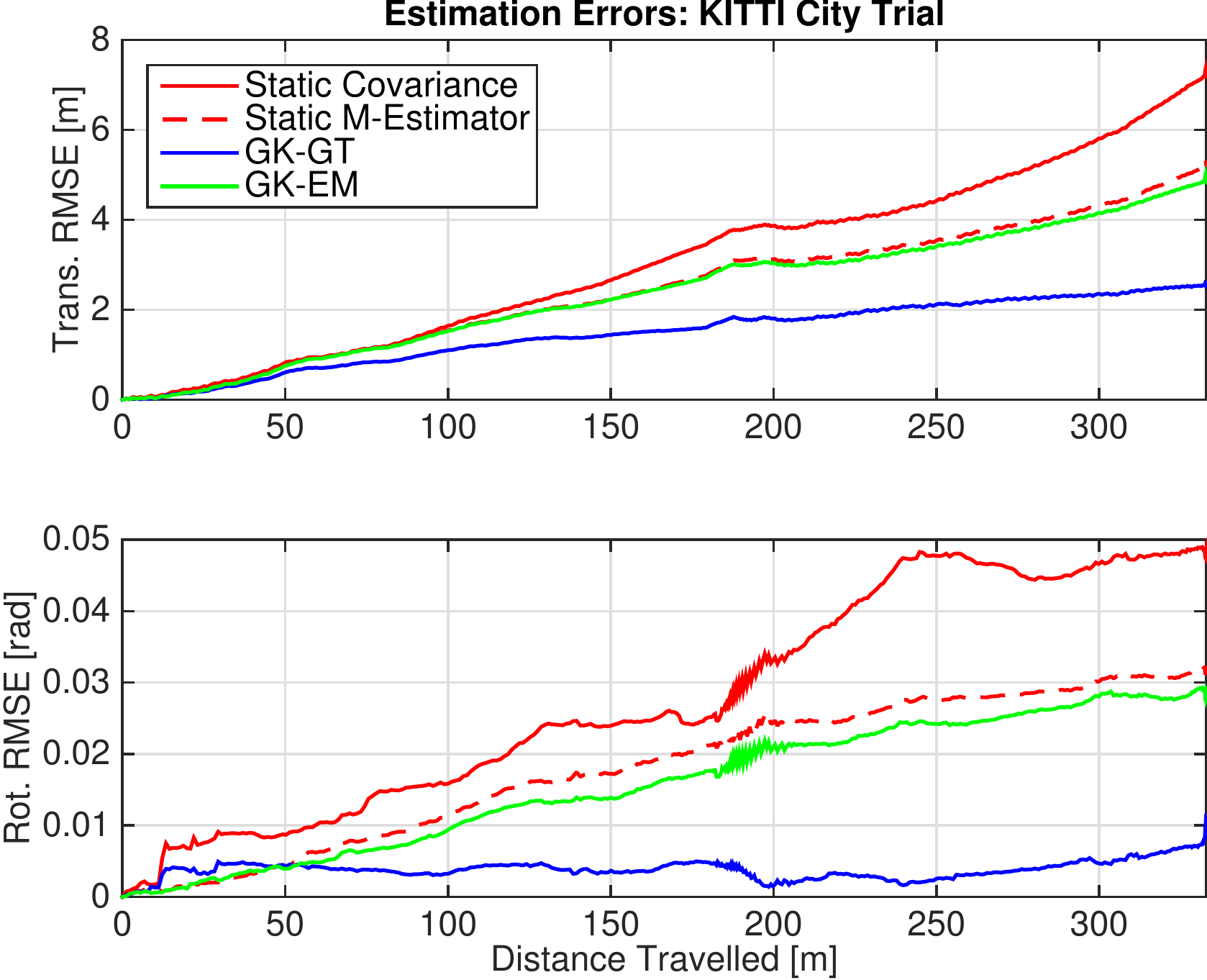}
    \caption{RMSE comparison of stereo odometry estimators evaluated on data from the city category in the KITTI dataset. See \Cref{table:armse_errors} for a quantitative summary.}
    \vspace{-0.4em}
    \label{fig:kitti_comparison1}
\end{figure}

Our prediction space consisted  of inertial magnitudes, high and low image frequency
coefficients, image entropy, pixel location, and estimated transform parameters.
The choice of predictors is motivated by the types of effects we wish to capture
(in this case: grassy self-similar textures, as well as shadows, and motion blur). For
a more detailed explanation of our choice of prediction space, see our previous work
\cite{peretroukhin2015PROBE}.

\begin{figure}
    \centering
    \includegraphics[width=0.45\textwidth]{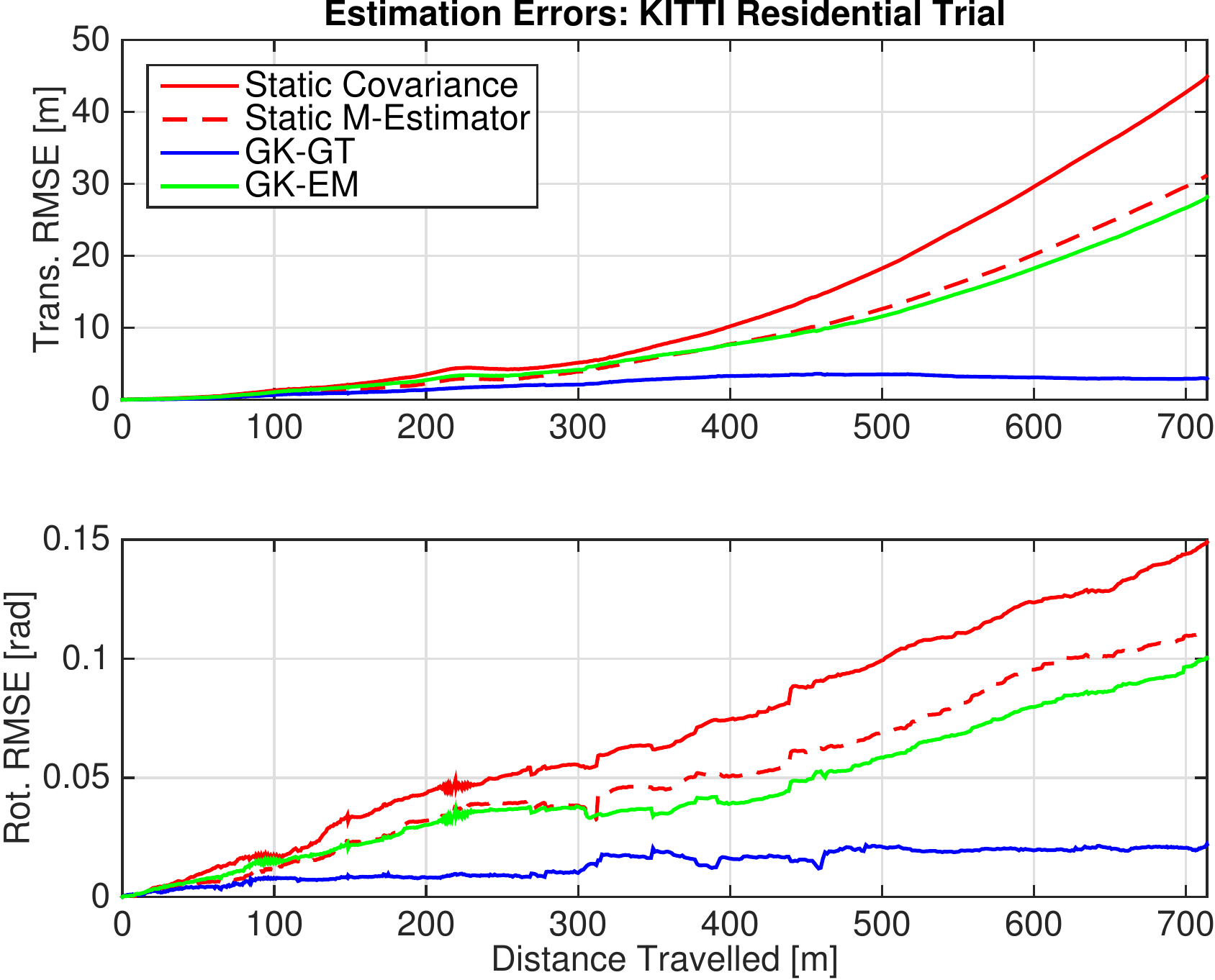}
    \caption{RMSE comparison of stereo odometry estimators evaluated on data from the residential category in the KITTI dataset. See \Cref{table:armse_errors} for a quantitative summary.}
    \vspace{-0.4em}
    \label{fig:kitti_comparison2}
\end{figure}

\begin{figure}
    \centering
    \includegraphics[width=0.45\textwidth]{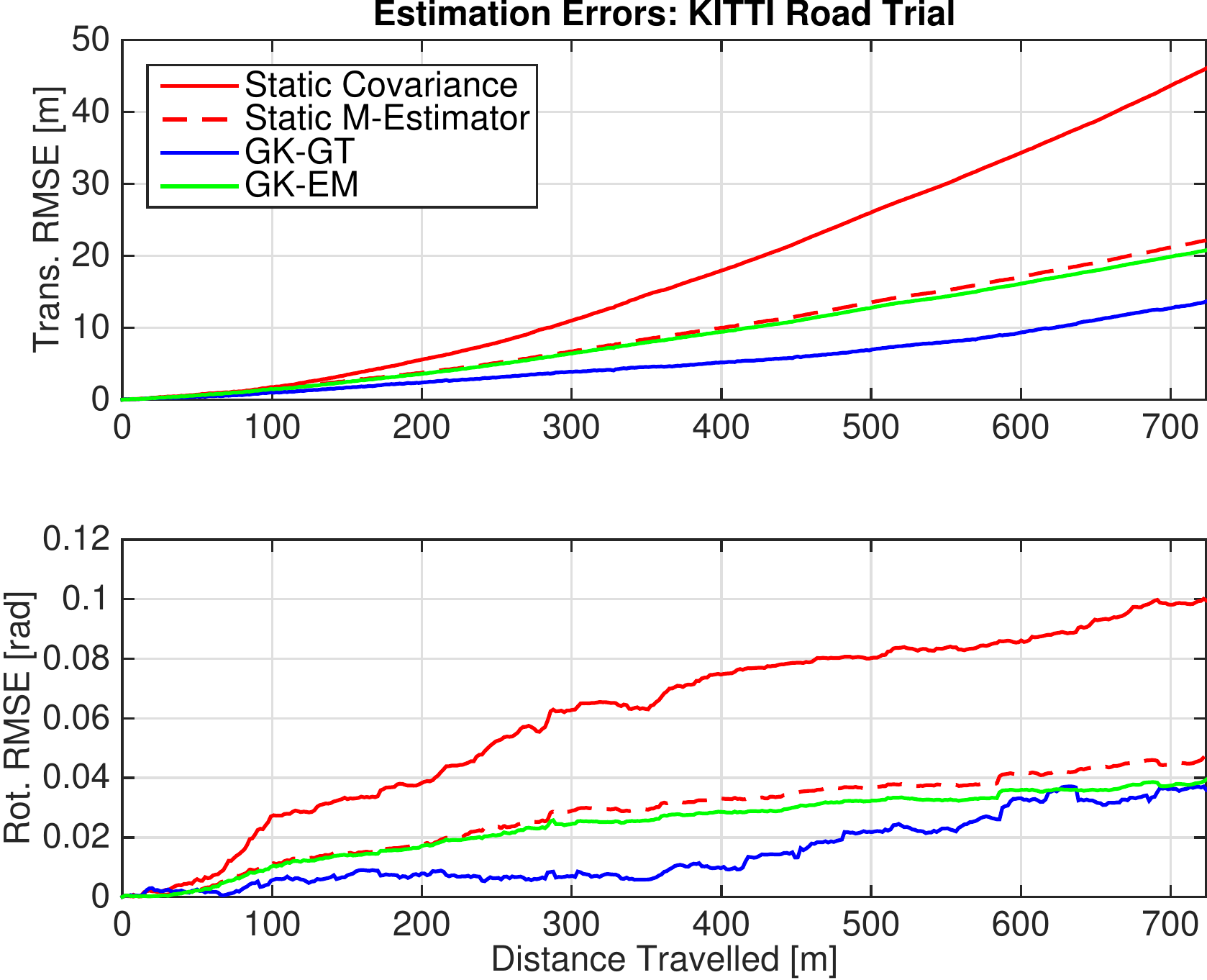}
    \caption{RMSE comparison of stereo odometry estimators evaluated on data from the road category in the KITTI dataset. See \Cref{table:armse_errors} for a quantitative summary.}
    \vspace{-0.4em}
    \label{fig:kitti_comparison3}
\end{figure}

\Cref{fig:kitti_comparison1,fig:kitti_comparison2,fig:kitti_comparison3} show
typical results; \Cref{table:armse_errors} presents a quantitative comparison.
PROBE GK-GT produced significant reductions in ARMSE, reducing translational ARMSE by
as much as 80\%. In contrast, GK-EM showed more modest improvements; this is
unlike our synthetic experiments, where both GK-EM and GK-GT achieved similar
performance. We are still actively exploring why this is the case; we note that although our simulated
data is drawn from a mixture of Gaussian distributions, the underlying
noise distribution for real data may be far more complex. With no ground truth, EM has to jointly optimize the camera poses and sensor uncertainty. It is unclear whether this is feasible in the general case with no ground truth information.

Further, we observe that the performance of PROBE-GK depends on the similarity
of the training data to the final test trials. A characteristic training dataset was important for consistent improvements on test trials.

\subsection{Experimental Dataset}

\begin{table*}
\centering
\caption{Comparison of average root mean squared errors (ARMSE) for rotational
  and translational components. Each trial is trained and tested from a
  particular category of raw data from the synthetic and KITTI datasets.}
\begin{tabular}{l c c c c c c c c c }
 & & \multicolumn{4}{c}{Trans. ARMSE [m]} & \multicolumn{4}{c}{Rot. ARMSE [rad]}  \\  \cline{3-6}  \cline{7-10} \T                                                                                    
 & Length [m] & Fixed Covar. & Static M-Estimator  & GK-GT & GK-EM & Fixed Covar. &  Static M-Estimator  & GK-GT & GK-EM  \\                         
\hline \T
Synthetic & 180 & 3.87 & 2.49 & 1.59 & 1.66 & 0.18 & 0.13 & 0.070 & 0.073 \\                                                                                                                
City & 332.9 & 3.84 & 2.99 & 1.69 & 2.87 & 0.032 & 0.021 & 0.0046 & 0.018 \\ 
Residential & 714.1 & 13.48 & 9.37 & 1.97 & 8.80 & 0.068 & 0.050 & 0.013 & 0.044 \\
Road & 723.8 & 17.69 & 9.38 & 5.24 & 8.87 & 0.060 & 0.027 & 0.015 & 0.024
\\ \hline                                                                                                                
\end{tabular}               \label{table:armse_errors}
\end{table*}

\begin{figure}
    \centering
    \hspace*{0.25cm}
    \includegraphics[width=0.45\textwidth]{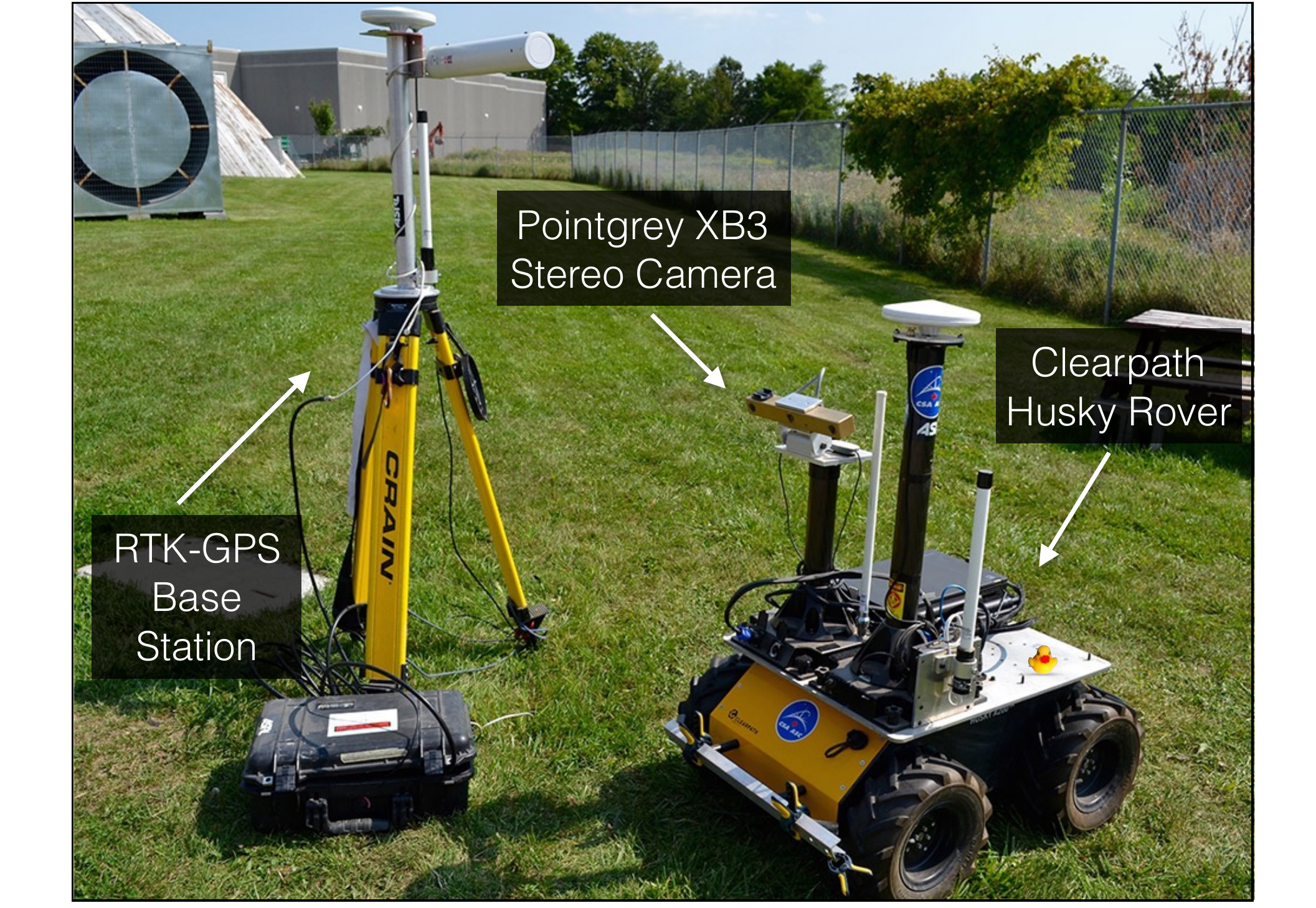}
    \caption{Our experimental apparatus: a Clearpath Husky rover outfitted with a PointGrey XB3 stereo camera and a differential GPS receiver and base station.}
      \vspace{-0.5em}
   	    \label{fig:experiments}
\end{figure}

\begin{figure}
    \centering
	\includegraphics[width=0.45\textwidth]{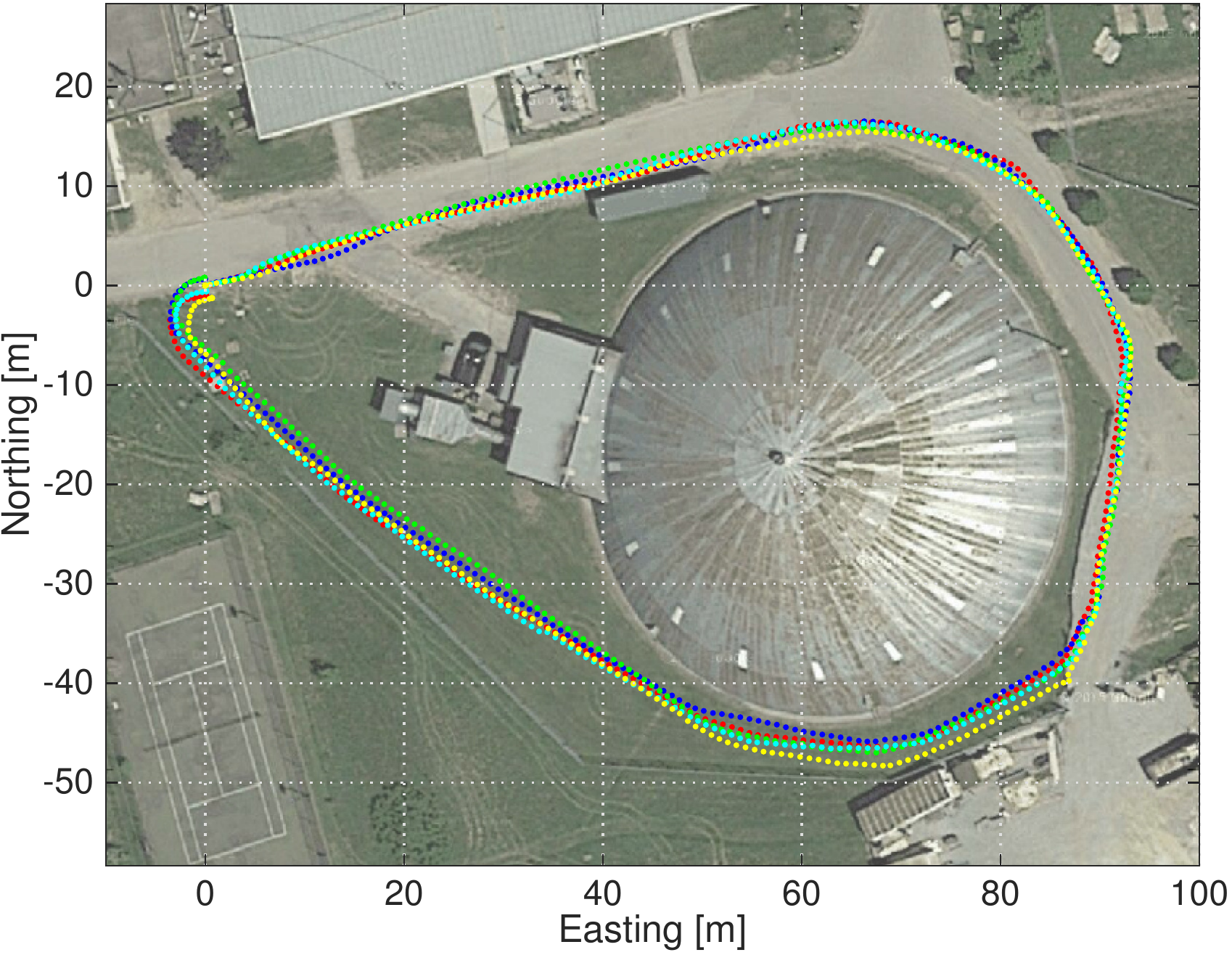}
    \caption{GPS ground truth for 5 experimental trials collected
      near the UTIAS Mars Dome. Each trial is approximately 250 m long.}
      \vspace{-0.5em}
    \label{fig:experiment_groundtruth}
\end{figure}

To further investigate the capability of our EM approach, we evaluated PROBE-GK on experimental data collected at the University of Toronto Institute for Aerospace Studies (UTIAS). For this experiment, we drove a Clearpath Husky rover outfitted with an Ashtech DG14 Differential GPS, and a PointGrey XB3 stereo camera around the MarsDome (an indoor Mars analog testing environment) at UTIAS (\Cref{fig:experiments}) for five trials of a similar path.  Each trial was approximately 250 m in length and we made an effort to align the start and end points of each loop. We used the wide baseline (25 cm) of the XB3 stereo camera to record the stereo images. The approximate trajectory for all 5 trials, as recorded by GPS, is shown in \Cref{fig:experiment_groundtruth}.  Note that the GPS data was not used during training, and only recorded for reference.

\begin{figure}
    \centering
    \includegraphics[width=0.45\textwidth]{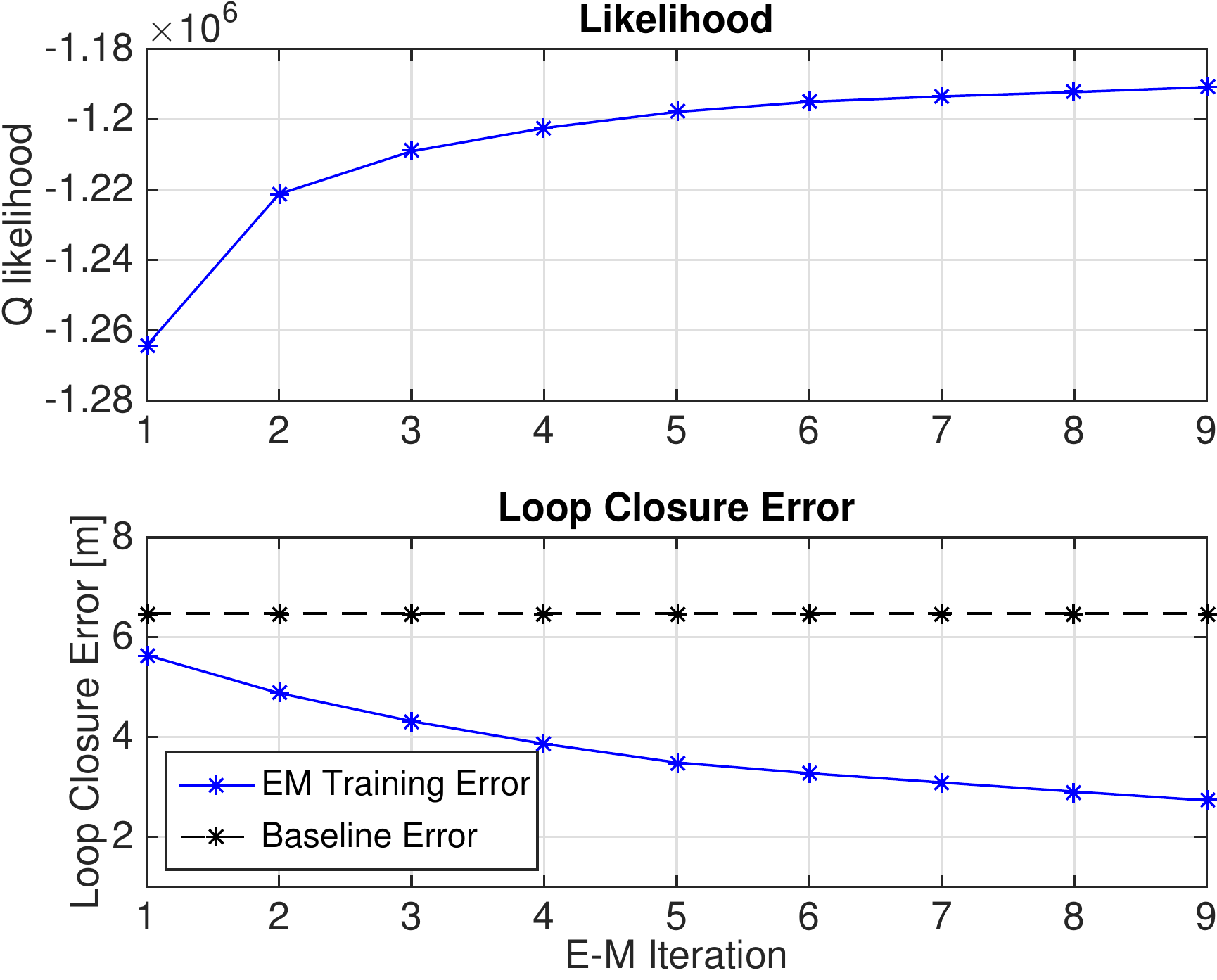}
    \caption{Training without ground truth using PROBE-GK-EM on a 250.2m path
      around the Mars Dome at UTIAS. The likelihood of the data increases with
      each iteration, and the loop closure error decreases, improving significantly from a baseline static M-estimator.}
      \vspace{-1em}
    \label{fig:experiments_trainingstats}
\end{figure}

For the prediction space in our experiments, we mimicked the KITTI experiments, omitting inertial magnitudes as no inertial data was available. We trained PROBE-GK without ground truth, using the Expectation Maximization approach. \Cref{fig:experiments_trainingstats} shows the likelihood and loop closure error as a function of EM iteration. 

The EM approach indeed produced significant error reductions on the training dataset after just a few iterations.  Although  it was trained with no ground truth information, our PROBE-GK model was used to produce significant reductions in the loop closure errors of the remaining 4 test trials. This reinforced our earlier hypothesis: the EM method works well when the training trajectory more closely resembles the test trials (as was the case in this experiment). \Cref{table:loop_closure_errors} lists the statistics for each test.

\begin{table}
\centering
\caption{Comparison of loop closure errors for 4 different experimental trials
  with and without a learned PROBE-GK-EM model.}
\begin{tabular}{ c  c  c  c }
     & & \multicolumn{2}{c}{Loop Closure Error [m]}  \\ \cline{3-4} \T
    Trial & Path Length [m] & PROBE-GK-EM & Static M-Estimator \\    
      \hline \T	
  2 & 250.3 & 3.88 & 8.07 \\
  3 & 250.5 & 3.07 & 6.64 \\
  4 & 205.4 & 2.81 & 7.57 \\
  5 & 249.9 & 2.34 & 7.75 \\ \hline
\end{tabular}
\label{table:loop_closure_errors}
\end{table}

\vspace{1em}
\section{Related Work}

There is a large and growing body of work on the problem of deriving accurate,
consistent state estimates from visual data.  Although our approach to noise
modelling is applicable in other domains, for simplicity we focus our attention
on the problem of inferring egomotion from features extracted from sequential
pairs of stereo images; see \cite{sunderhauf2007stereo} for a survey of
techniques. The spectrum of alternative approaches to visual state estimation
include monocular techniques, which may be feature-based
\cite{scaramuzza2011visual}, direct \cite{irani2000direct}, or semi-direct
\cite{forster2014svo}. 

Apart from simply rejecting outliers, a number of recent approaches attempt to
select the optimal set of features to produce an accurate localization estimate
from tracked visual features. For example, \cite{Tsotsos2015} amend Random
Sample Consensus (RANSAC) with statistical hypothesis testing to ensure that tracked visual features have normally distributed residuals before including them in
the estimator. Unlike our predictive approach, their technique relies on the availability of feature tracks, and requires scene overlap to work continuously. In a different
approach, \cite{Zhang2015} choose an optimally observable feature subset for a
monocular SLAM pipeline by selecting features with the highest \textit{informativeness} - a measure calculated based on the observability of the SLAM subsystem. Observability, however, is governed by the 3D location of the features, and therefore cannot predict systematic feature degradation due to environmental or sensor-based effects. In contrast, PROBE-GK can leverage prior data to learn such effects and map them to predicted uncertainty on visual observations, optimally weighting the
contribution of each observation to the final state estimate.


\section{Conclusion}
The method presented in this paper applies the technique of
generalized kernel estimation to improve on the uncorrelated and static Gaussian error
models typically employed in stereo odometry. By inferring a more accurate
noise model given past sensory experience, we can reduce the tracking error of
a sequence of estimates and improve the robustness of our estimator, even when
the training data does not have associated ground truth. 
Our method has the advantage of having relatively few tuning parameters,
meaning it can be applied to new problems with very little user intervention. We
do rely on the availability of a good set of predictors, and have found that for
problems of interest finding a good set is not difficult;  a principled choice of an optimal set of predictors, however, remains an interesting open problem. 

\balance

Although our experiments demonstrate utility only in the context of sequential
maximum likelihood estimation on stereo vision data, we believe the model presented
here can be applied to a more general class of filter or factor-based estimation
algorithms, as well as to a more general class of sensors. In future work, we plan to investigate the
applicability of our method to problems of simultaneous localization and
mapping, explore the possibility of learning the predictive model online (obviating the need for training data), and examine more principled approaches to selecting an informative prediction space.

\bibliographystyle{IEEEtran}
\bibliography{peretroukhin_icra16.bib}

\end{document}

%% file: variables.tex
\NewDocumentCommand\ArgMax{m}{\underset{#1}{\text{arg\,max}}\,}
\NewDocumentCommand\ArgMin{m}{\underset{#1}{\text{arg\,min}}\,}

\NewDocumentCommand\NaturalNumbers{oo}{%
    \IfNoValueTF{#2}
    {
        \IfNoValueTF{#1}
        {\ensuremath{\mathbb N}}
        {\ensuremath{\mathbb N^{#1}}}
    }
    { \ensuremath{\mathbb N^{#1\times #2}}}
}
\NewDocumentCommand\Integers{oo}{%
    \IfNoValueTF{#2}
    {
        \IfNoValueTF{#1}
        {\ensuremath{\mathbb Z}}
        {\ensuremath{\mathbb Z^{#1}}}
    }
    { \ensuremath{\mathbb Z^{#1\times #2}}}
}
\NewDocumentCommand\RationalNumbers{oo}{%
    \IfNoValueTF{#2}
    {
        \IfNoValueTF{#1}
        {\ensuremath{\mathbb Q}}
        {\ensuremath{\mathbb Q^{#1}}}
    }
    { \ensuremath{\mathbb Q^{#1\times #2}}}
}
\NewDocumentCommand\RealNumbers{oo}{%
    \IfNoValueTF{#2}
    {
        \IfNoValueTF{#1}
        {\ensuremath{\mathbb R}}
        {\ensuremath{\mathbb R^{#1}}}
    }
    { \ensuremath{\mathbb R^{#1\times #2}}}
}
\NewDocumentCommand\ComplexNumbers{oo}{%
    \IfNoValueTF{#2}
    {
        \IfNoValueTF{#1}
        {\ensuremath{\mathbb C}}
        {\ensuremath{\mathbb C^{#1}}}
    }
    { \ensuremath{\mathbb C^{#1\times #2}}}
}
\NewDocumentCommand\HomogeneousNumbers{oo}{%
    \IfNoValueTF{#2}
    {
        \IfNoValueTF{#1}
        {\ensuremath{\mathbb P}}
        {\ensuremath{\mathbb P^{#1}}}
    }
    { \ensuremath{\mathbb P^{#1\times #2}}}
}

\def\Vec#1{\!\!\hbox{$#1$\kern-0.38em\lower0.85em\hbox{$\vec{}\,$}}\,}%
\newcommand{\bbm}{\begin{bmatrix}}
\newcommand{\ebm}{\end{bmatrix}}
\DeclareMathAlphabet{\mbf}{OT1}{ptm}{b}{n}

\NewDocumentCommand\Transpose{m}{{#1}^\top}
\NewDocumentCommand\Vector{m}{ \ensuremath{ \mathbf{\boldsymbol{#1}} } } 
\NewDocumentCommand\Matrix{m}{ \ensuremath{ \mathbf{#1} } } 

\NewDocumentCommand\Trace{m}{\ensuremath{\mathrm{tr}\left(#1\right)}}

\NewDocumentCommand\Identity{o}{%
    \IfNoValueTF{#1}
    {\ensuremath{\mathbbm 1}}
    {\ensuremath{\mathbbm 1_{#1}}}
}

\NewDocumentCommand\Zero{o}{%
    \IfNoValueTF{#1}
    {\ensuremath{\mathbf 0}}
    {\ensuremath{\mathbf 0_{#1}}}
}

\NewDocumentCommand\HomogeneousPoint{mm}{\boldsymbol{p}_{#1,#2}}
\nomenclature{$\HomogeneousPoint{i}{t}\in\HomogeneousNumbers[3]$}{Location of visual landmark i during frame t in homogeneous coordinates.}%

\NewDocumentCommand\VisualLandmark{mm}{\Vector{z}_{#1,#2}}
\nomenclature{$\VisualLandmark{i}{t}\in\RealNumbers[3]$}{Location of visual
  landmark $i$ in the global coordinate frame during image frame $t$}%

\NewDocumentCommand\CameraPose{m}{\Vector{x}_{#1}}
\nomenclature{$\CameraPose{t}\in\text{SE}(3)$}{Camera pose at frame $t$}%

\NewDocumentCommand\ImageLandmark{mm}{\Vector{y}_{#1,#2}}

\NewDocumentCommand\TargetImageLandmark{o}{
    \IfNoValueTF{#1}
    {\Vector{y}_{*}}
    {\Vector{y}_{*,#1}}
  }
\nomenclature{$\ImageLandmark{i}{t}\in\RealNumbers[4]$}{Location
  of visual landmark $i$ during frame $t$ in image coordinates
  for each stereo camera.}%

\NewDocumentCommand\ImageLandmarkCovariance{mm}{\Matrix{R}_{#1,#2}}
\NewDocumentCommand\Covariance{}{\Matrix{R}}

\NewDocumentCommand\TargetImageLandmarkCovariance{}{\Matrix{R}_*}
\nomenclature{$\ImageLandmarkCovariance{i}{t}\in\RealNumbers[4\times
  4]\succ 0$}{Location
  of visual landmark $i$ during frame $t$ in image coordinates
  for each stereo camera.}%

\NewDocumentCommand\Predictor{mm}{\Vector{\phi}_{#1,#2}}
\nomenclature{$\Predictor{i}{t}$}{Predictor features for image-space
  landmark $i$ in frame $t$}%

\NewDocumentCommand\TargetPredictor{}{\Vector{\phi}_*}
\nomenclature{$\Predictor{i}{t}$}{Predictor features for image-space
  landmark $i$ in frame $t$}%

\NewDocumentCommand\Transform{}{\mathcal{T}}

\nomenclature{$\Transform_t=\Transform(\CameraPose{t})$}{The operator which
  applies a frame transform to a vector, so that $\Transform_t
  \VisualLandmark{i}{t}$ is the location of visual landmark $i$ relative to the
  camera coordinate frame.}

\NewDocumentCommand\ProjectionFunction{}{f}
\nomenclature{$\ProjectionFunction:
  \RealNumbers[3]\to\RealNumbers[4]$}{Function mapping a visual landmark in
  the camera coordinate frame to image coordinates, so that
  $\ProjectionFunction( \VisualLandmark{i}{t}) =
  \ImageLandmark{i}{t}$. This function is invertible, so that
  $\ProjectionFunction^{-1}( \ImageLandmark{i}{t}) =
  \VisualLandmark{i}{t}$.}%

\NewDocumentCommand\NormalDistribution{}{\mathcal{N}}
\nomenclature{$\NormalDistribution\left(\Vector{x};
    \Vector{\mu},\Matrix{\Sigma}\right)$}{A multivariate normal distribution
  with mean $\Vector{\mu}\in\RealNumbers[d]$ and covariance
  $\Matrix{\Sigma}\in\RealNumbers[d\times d]\succ\Matrix{0}$: 
  \begin{equation*}
    \NormalDistribution\left(\Vector{x};\Vector{\mu},\Matrix{\Sigma}\right)
    = (2\pi)^{-\frac{d}{2}}\vert\Matrix{\Sigma}\vert^{-\frac{1}{2}}
    \exp\left( -\frac{1}{2}
      \Transpose{(\Vector{x}-\Vector{\mu})} \Matrix{\Sigma}^{-1}
      (\Vector{x}-\Vector{\mu}) \right)
  \end{equation*}  
}

\NewDocumentCommand\InverseWishartDistribution{}{\text{IW}}
\nomenclature{$\InverseWishartDistribution\left(\Matrix{R};
    \Matrix{\Psi},\nu\right)$}{An inverse Wishart distribution 
  with scale matrix $\Matrix{\Psi}\in\RealNumbers[d\times d]\succ\Matrix{0}$
  and degrees of freedom $\nu>d-1$.  
  \begin{equation*}
    \InverseWishartDistribution\left(\Matrix{R}; \Matrix{\Psi}, \nu\right)
    = \frac{\vert\Matrix{\Psi}\vert^{\frac{\nu}{2}}}{2^{\frac{\nu
          d}{2}}\Gamma_d(\frac{\nu}{2})}
    \vert\Matrix{R}\vert^{-\frac{\nu+d+1}{2}}
    \exp\left( -\frac{1}{2}
      \Trace{\Matrix{\Psi}\Matrix{R}^{-1}}\right)
  \end{equation*}  
}

\NewDocumentCommand\StudentTDistribution{}{\text{t}}
\nomenclature{$\StudentTDistribution_{\nu}\left(\Vector{x}; \Vector{\mu},
    \Matrix{\Sigma}\right)$}{ A multivariate student's $t$-distributon with mean
  $\Vector{\mu}\in\RealNumbers[d]$, covariance $\Matrix{\Sigma} \in
  \RealNumbers[d\times d] \succ\Matrix{0}$, and degrees of freedom $\nu>0$.  
  \begin{equation*} 
    \StudentTDistribution_{\nu}\left(\Vector{x}; \Vector{\mu},
      \Matrix{\Sigma}\right) =
    \frac{\Gamma(\frac{\nu+d}{2})}{\Gamma(\frac{\nu}{2})}
    \vert\Matrix{\Sigma}\vert^{-\frac{1}{2}}(\nu\pi)^{-\frac{d}{2}}
    \left(1+\frac{1}{\nu}
      \Transpose{(\Vector{x}-\Vector{\mu})} \Matrix{\Sigma}^{-1}
      (\Vector{x}-\Vector{\mu}) \right)^{-\frac{\nu+d}{2}}
  \end{equation*}  
}